\documentclass[10pt,twocolumn,letterpaper]{article}

\usepackage[pagenumbers]{cvpr} % To force page numbers, e.g. for an arXiv version

\usepackage{graphicx}
\usepackage{amsmath}
\usepackage{amssymb}
\usepackage{booktabs}
\usepackage{multirow}
\usepackage[pagebackref=true,breaklinks=true,colorlinks,bookmarks=false]{hyperref}

\usepackage[capitalize]{cleveref}
\crefname{section}{Sec.}{Secs.}
\Crefname{section}{Section}{Sections}
\Crefname{table}{Table}{Tables}
\crefname{table}{Tab.}{Tabs.}

\begin{document}

\title{GPS-GLASS: Learning Nighttime Semantic Segmentation Using Daytime Video and GPS data} 

\author{Hongjae Lee, Changwoo Han, Jun-Sang Yoo, Seung-Won Jung\thanks{Corresponding author}\\
	Department of Electrical Engineering, Korea University\\
	{\tt\small \{jimmy9704, hcwoo329, junsang7777, swjung83\}@korea.ac.kr}
}

\maketitle

%%%%%%%%% ABSTRACT
\begin{abstract}
    Semantic segmentation for autonomous driving should be robust against various in-the-wild environments. Nighttime semantic segmentation is especially challenging due to a lack of annotated nighttime images and a large domain gap from daytime images with sufficient annotation. In this paper, we propose a novel GPS-based training framework for nighttime semantic segmentation. Given GPS-aligned pairs of daytime and nighttime images, we perform cross-domain correspondence matching to obtain pixel-level pseudo supervision. Moreover, we conduct flow estimation between daytime video frames and apply GPS-based scaling to acquire another pixel-level pseudo supervision. Using these pseudo supervisions with a confidence map, we train a nighttime semantic segmentation network without any annotation from nighttime images. Experimental results demonstrate the effectiveness of the proposed method on several nighttime semantic segmentation datasets. Our source code is available at \url{https://github.com/jimmy9704/GPS-GLASS}.
\end{abstract}

%%%%%%%%% BODY TEXT
\section{Introduction}
\begin{figure}[!t]
  \centering
  \includegraphics[width=0.9\linewidth]{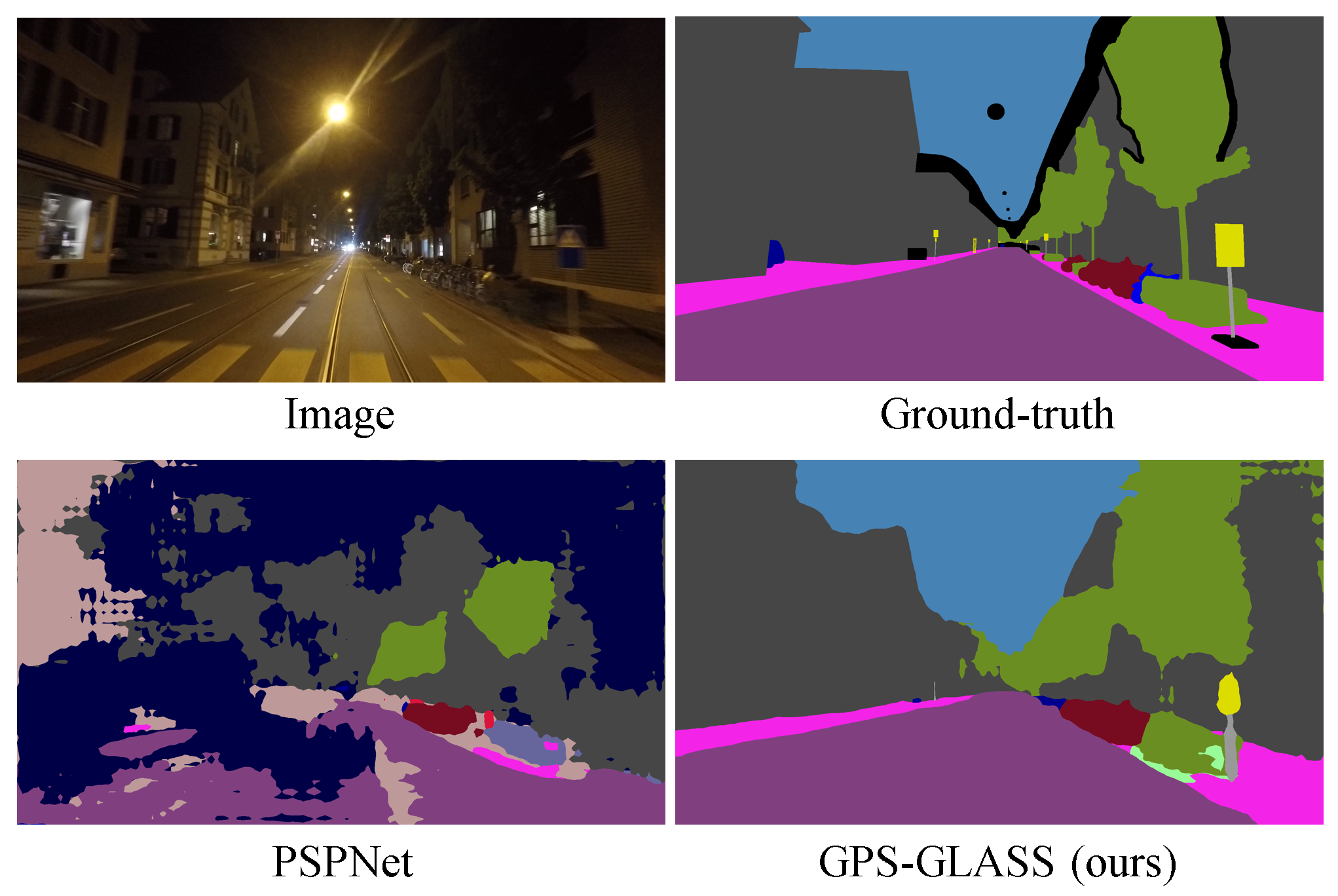}
  \caption{Visual comparison of the nighttime semantic segmentation
results between the PSPNet~\cite{zhao2017pyramid} without domain adaptation and our proposed GPS-GLASS.}
  \label{fig:intro}
\end{figure}

{S}{emantic} segmentation, which classifies each pixel of an image into a semantic class, is a fundamental problem in computer vision and has been widely used in various applications, including autonomous driving, robotic navigation, and medical imaging. In particular, for autonomous driving applications, it is necessary to design a segmentation method that is robust against domain changes such as illumination and weather changes. In order to design such a method, especially with convolutional neural networks (CNNs), a large amount of pixel-level annotated data is required for supervised learning. However, acquiring pixel-level annotation in poor illumination environments such as nighttime is very challenging beyond the cost of annotations. Therefore, most semantic segmentation datasets focus primarily on daytime environments~\cite{Cordts2016Cityscapes,Geiger2013kitti}, but a semantic segmentation model trained on these datasets fails in nighttime semantic segmentation, as shown in Fig.~\ref{fig:intro}. Although some datasets~\cite{dai2018DMAda,yu2020bdd100k} provide nighttime image annotations, their quantity and quality are insufficient to be used for semantic segmentation network training. In this paper, we propose a training methodology for nighttime image semantic segmentation networks without requiring pixel-level annotation of nighttime scenes. 

Several methods have been developed to adapt daytime segmentation networks to nighttime scenes without using annotated nighttime images. For example, the twilight domain between daytime and nighttime has been introduced for gradual domain adaptation
~\cite{dai2018DMAda,sakaridis2019guided,sakaridis2020map-guided}. Image translation has also been attempted to obtain synthetic annotations of nighttime images that can help train semantic segmentation networks~\cite{romera2019bridging,sun2019see}. However, these methods require additional training data in the twilight domain or several pre-processing stages. Several recent methods~\cite{wu2021dannet,xu2021cdada} have presented pseudo-supervised loss terms using coarsely aligned daytime and nighttime image pairs. These recent methods require neither additional domain data nor pre-processing stages, but they have not attempted to align daytime and nighttime image pairs precisely. 

\begin{figure*}[t]
  \includegraphics[width=\textwidth]{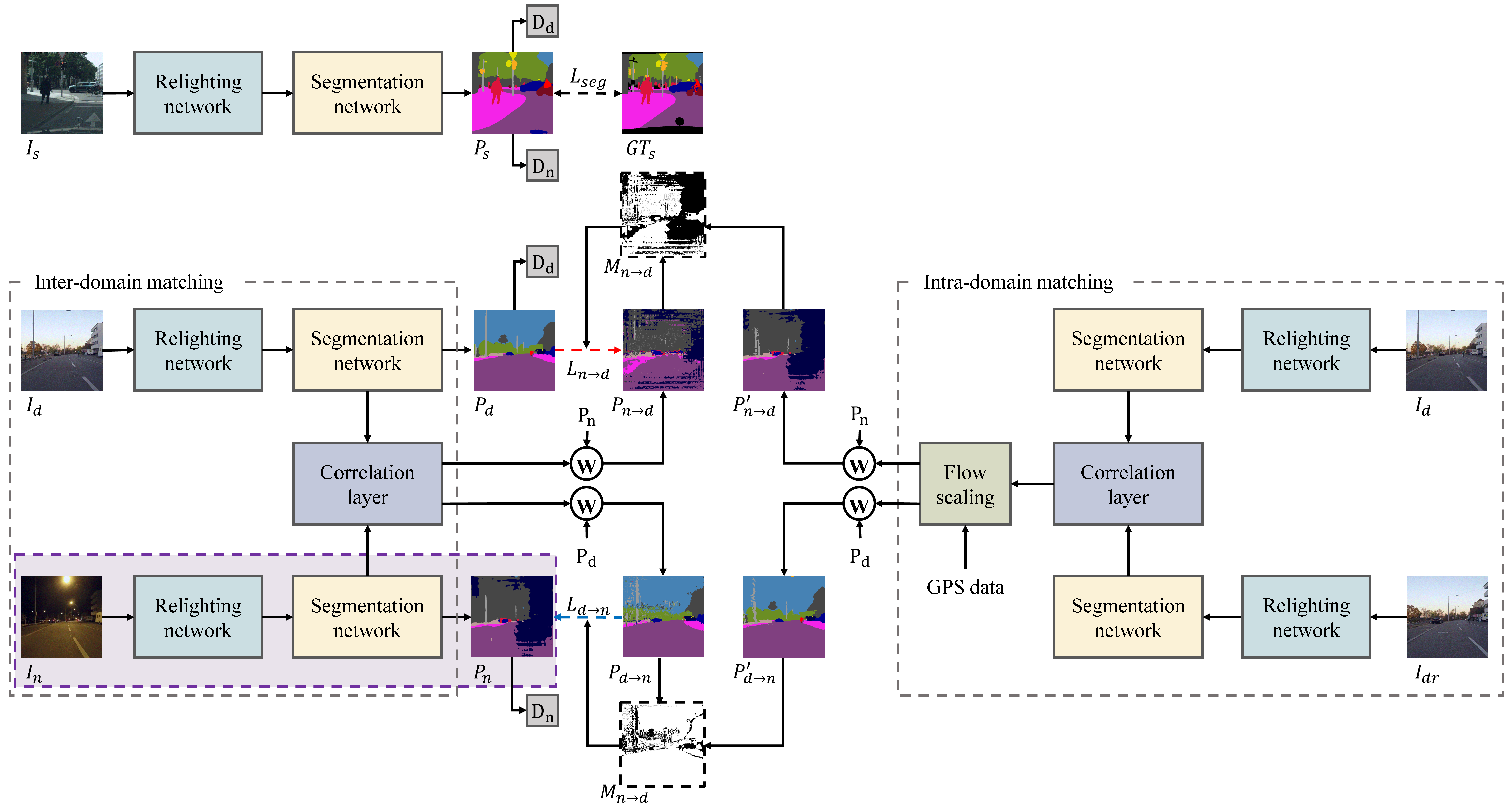}
  \caption{The overview of the proposed GPS-GLASS. The same colored networks share weights, and the correlation layer has no weights that require training. \textcircled{w} represents the backward warping operator, and the red and blue dotted arrows indicate supervision by the ground-truth and pseudo-labels, respectively. Only the networks inside the purple box are used at the inference stage.}
  \label{fig:overview}
\end{figure*}

In this paper, we present a novel Global Positioning System (GPS)-Guided Learning Approach for nighttime Semantic Segmentation (GPS-GLASS), as illustrated in Fig.~\ref{fig:overview}. Similar to DANNet~\cite{wu2021dannet}, GPS-GLASS uses image relighting and semantic segmentation modules and two discriminators for the daytime and nighttime domains. Unlike DANNet, GPS-GLASS extracts image features obtained during the segmentation process to estimate the correspondence from the daytime to nighttime and vice versa. Moreover, observing that nighttime images are located between daytime image frames, GPS-GLASS applies intra-domain correspondence matching to daytime image frames and performs GPS-based flow scaling. From these inter-domain and intra-domain correspondences, we construct pseudo-labels for training a nighttime semantic segmentation network. In addition, due to the cross-domain correspondence matching the proposed GPS-GLASS well generalized to both daytime and nighttime.

Our contributions are summarized as follows:
\begin{itemize}
\item{We introduce a framework called GPS-GLASS that performs inter-domain correspondence matching to construct a pseudo-label for training a nighttime semantic segmentation network.}
\item {We propose to perform intra-domain correspondence matching using daytime video frames and scale the estimated flow field using GPS data, yielding another pseudo-label.}
\item{By combining the two pseudo-labels with a confidence map, GPS-GLASS shows state-of-the-art performance on several nighttime image datasets. Ablation studies also verify the effectiveness of each component of GPS-GLASS.}
\end{itemize}

\section{Related Work}
\subsection{Unsupervised Domain Adaptation for Nighttime Semantic Segmentation}
Supervised training of semantic segmentation networks requires pixel-level annotation, which is laborious and time-consuming to obtain. Because ground-truth annotation is publicly available only for some limited domains, e.g., Cityscapes~\cite{Cordts2016Cityscapes} for daytime road scenes and GTA5~\cite{richter2016GTA} for synthetic scenes, unsupervised domain adaptation (UDA) has received significant interest. There have been several approaches~\cite{tsai2019domain,Lee2021content} to achieve the goal of UDA for semantic segmentation. However, these approaches are focused on reducing the domain gap between synthetic and real. 

The existing datasets developed for semantic segmentation of road scenes are biased toward daytime scenes~\cite{Cordts2016Cityscapes,Geiger2013kitti}, segmentation networks trained without considering UDA tend to fail in handling nighttime scenes. Consequently, efforts have been made to reduce the domain gap between daytime and nighttime scenes.

Motivated by the widely studied image style transfer~\cite{jeong2021memory,jiang2020tsit,liu2017unit}, one can try translating daytime scenes to nighttime scenes. Earlier studies along this direction~\cite{romera2019bridging,sun2019see} were overly interested in the auxiliary task of the image style transfer rather than the main semantic segmentation task. Sakaridis~\etal~\cite{sakaridis2019guided} constructed the \textit{Dark Zurich} dataset, which contains daytime and nighttime images that are coarsely matched with GPS information. These daytime and nighttime image pairs are helpful in guiding the semantic segmentation of nighttime scenes, resulting in many follow-up studies, including their guided curriculum model adaptation (GCMA). Sakaridis~\etal~\cite{sakaridis2020map-guided} proposed an improved version of GCMA that uses depth and camera pose information. Wu~\etal~\cite{wu2021dannet} proposed a multi-target domain adaptation network for nighttime semantic segmentation via adversarial learning. Xu~\etal~\cite{xu2021cdada} proposed nighttime domain gradual self-training and patch-level prediction guidance methods. Gao~\etal~\cite{gao2022ccdistill} proposed a correlation distillation approach for cross-domain between synthetic and real nighttime.

However, the above methods~\cite{sakaridis2019guided,sakaridis2020map-guided,wu2021dannet,xu2021cdada} have not attempted to precisely align daytime and nighttime images because such an alignment can even be a more difficult task than semantic segmentation. Xu~\etal~\cite{wu2021dania} aligned the daytime and nighttime images using an additional optical flow estimation network. However, the above method requires additional datasets and training stages for the optical flow estimation network. We notice that the invaluable information of GPS is obtainable when constructing datasets such as Dark Zurich. Consequently, in this paper, we propose to use the GPS information of daytime and nighttime images to guide the correspondence matching for nighttime semantic segmentation network training.

\subsection{Optical Flow and Correspondence Matching}
Various studies are being conducted to find a matching point between two images e.g., stereo matching~\cite{5995372,6751115,6888475}, optical flow estimation~\cite{dosovitskiy2015flownet,ilg2017flownet}, and semantic correspondence~\cite{lee2020learning-sfnet,lee2019sfnet}. In the case of learning-based optical flow estimation approaches, Dosovitskiy~\etal~\cite{dosovitskiy2015flownet} introduced an end-to-end optical flow estimation method with CNNs. Ranjan~\etal~\cite{ranjan2017optical} proposed a spatial pyramid network that predicts flow in a coarse-to-fine manner. Sun~\etal~\cite{sun2018pwc} proposed a method of warping the spatial feature pyramid and calculating the cost volume from the warped features. Teed~\etal~\cite{teed2020raft} proposed a recurrent unit for gradual flow refinement, demonstrating high performance with fewer network parameters. Recent transformer~\cite{vaswani2017transformer}-based optical flow models~\cite{jiang2021GMA,xu2021high,xu2021gmflow} further improved the optical flow estimation performance.

Although the above methods have contributed to the development of optical flow estimation technology, they are still suffering from correspondence matching between different domains. Recently, Zhou~\etal~\cite{zhou2020cross-weather} estimated optical flow to match and align two images captured under different weather conditions. Zhang~\etal~\cite{zhang2020cross} proposed a method to train an image translation network by warping images from different domains. Lee~\etal~\cite{lee2020learning-sfnet,lee2019sfnet} introduced a model called SFNet that predicts bidirectional correspondence between different instances of the same object or scene category. Inspired by the superior performance of SFNet, we propose a dense corresponding matching method in different domains for nighttime semantic segmentation. A simple correlation layer trains the segmentation network to extract features that are invariant to the domain gap, e.g., between daytime and nighttime.

\section{Proposed Methods}
\subsection{Framework Overview}
Our method involves a source domain $S$ and two target domains $T_d$ and $T_n$, where $S$, $T_d$, and $T_n$ correspond to Cityscapes (daytime)~\cite{Cordts2016Cityscapes}, Dark Zurich-D (daytime), and Dark Zurich-N (nighttime)~\cite{sakaridis2019guided} datasets in our case study, respectively. Note that only the source domain has ground-truth segmentation labels, and the two target domains are coarsely paired according to GPS locations. As shown in Fig.~\ref{fig:overview}, our GPS-GLASS consists of a single weight-sharing relighting network ($G_R$), a single weight-sharing semantic segmentation network ($G_S$), and two discriminators ($D_d$ and $D_n$), where we used the same architecture of DANNet~\cite{wu2021dannet} for these network components. 
Let $I_s$, $I_n$, and $I_d$ denote image samples corresponding to $S$, $T_d$, and $T_n$, respectively. These images are fed to $G_R$ to make $G_S$ less sensitive to illumination changes~\cite{wu2021dannet}. The segmentation results are obtained as ${P_s} = {G_S}\left( {{G_R}\left( {{I_s}} \right)} \right)$, ${P_d} = {G_S}\left( {{G_R}\left( {{I_d}} \right)} \right)$, and ${P_n} = {G_S}\left( {{G_R}\left( {{I_n}} \right)} \right)$. Only ${P_s}$ has its corresponding ground-truth segmentation labels ${P_s^*}$, and the other two results ${P_d}$ and ${P_n}$ are supervised by the pseudo-label.

Specifically, the proposed training framework called GPS-GLASS obtains the pseudo-label by estimating dense correspondence between the daytime and nighttime images, where the correlation layer is applied to the intermediate features of the segmentation network. In addition, since the dense correspondence between daytime and nighttime images can be inaccurate, GPS-GLASS obtains another pseudo-label by estimating dense correspondence between the daytime images and applying GPS-based flow scaling. By using the two different sources for acquiring the pseudo-label, GPS-GLASS trains the nighttime semantic segmentation network without any annotation from nighttime images. The details of GPS-GLASS will be explained in the following subsections. 

\subsection{GPS-guided Learning Approach}

 \subsubsection{Correspondence matching using inter-domain}
\begin{figure}[t]
  \centering
  \includegraphics[width=\linewidth]{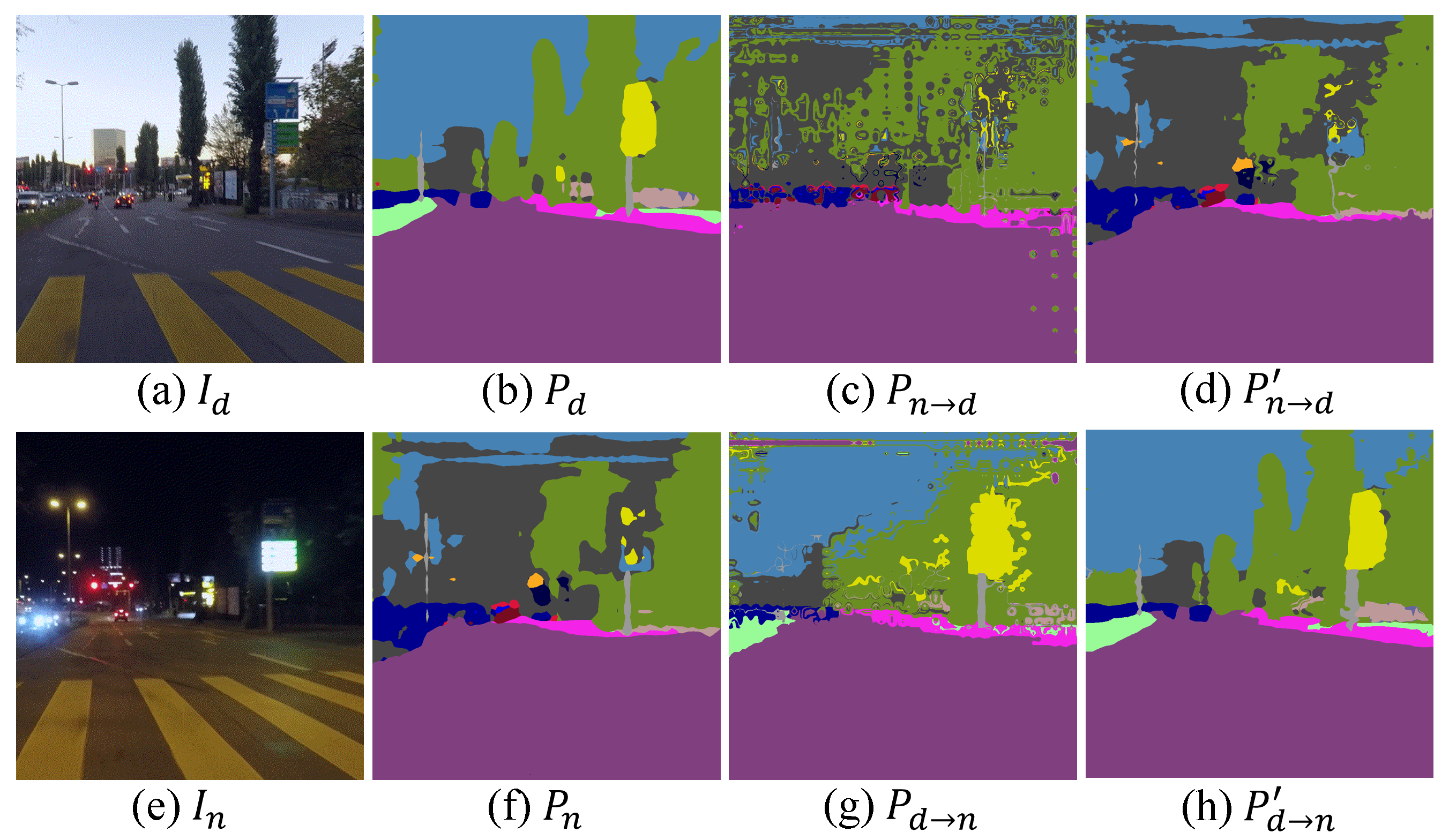}
  \caption{Examples of the segmentation results and warped pseudo-labels obtained during training. (a) and (e) are the input images, and (b) and (f) are the corresponding segmentation results. 
(c), (g) and (d), (h) are the results using the corresponding matching in the inter and intra domains, respectively.}
  \label{fig:flow}
\end{figure}

Our key idea is to align ${P_d}$ and ${P_n}$ such that the aligned segmentation result can be used as the pseudo-label. To this end, inspired by SFNet~\cite{lee2019sfnet}, the correlation layer is adopted to compute the dense correspondence of the image features between two different domains. A simple correlation layer without trainable parameters allows the segmentation network to extract features that are robust to domain changes, such as daytime and nighttime. Let ${f^d} = \left\{ {f_l^d,f_g^d} \right\}$ be the set of the local and global features extracted from the semantic segmentation network for the input $I_d$. In the case of PSPNet~\cite{zhao2017pyramid}, which is our chosen architecture for semantic segmentation, $f_l^d$ and $f_g^d$ are extracted before and after passing through the PSPmodule of PSPNet, respectively, and have the same dimension of $H \times W \times D$. ${f^n} = \left\{ {f_l^n,f_g^n} \right\}$ is extracted similarly from the semantic segmentation network for the input $I_n$. Then, the correlation layer computes the correlation between $f^d$ and $f^n$ as follows:
 \begin{equation}
\label{eqn: matching score}
{c_x}\left( {{\bf{p}},{\bf{q}}} \right) = {\left( {\frac{{f_x^d\left( {\bf{p}} \right)}}{{\left\| {f_x^d\left( {\bf{p}} \right)} \right\|}}} \right)^\top}\left( {\frac{{f_x^n\left( {\bf{q}} \right)}}{{\left\| {f_x^n\left( {\bf{q}} \right)} \right\|}}} \right),x \in \left\{ {l,g} \right\},
%c(\mathbf{p},\mathbf{q}) = f^{d}(\mathbf{p})^{\top}f^{n}(\mathbf{q}),
\end{equation}
where $\top$ is the transpose operator, $\left\|  \cdot  \right\|$ measures L2 norm, $\bf{p}$ and $\bf{q}$ represent 2D coordinates, and ${f_x^d\left( {\bf{p}} \right)}$ and ${f_x^n\left( {\bf{p}} \right)}$ are the $D$-dimensional vectors at $\bf{p}$ and $\bf{q}$, respectively. We combine the correlation volumes obtained from the local and global features by $c = {c_l} \odot {c_g}$, where $\odot$ represents element-wise multiplication. Instead of using the standard argmax function to obtain the correspondence from $c$, we use soft-argmax~\cite{honari2018improving,kendall2017end} to allow backpropagation through the correlation layer as follows:
 \begin{equation}
\label{eqn: softmax}
c'\left( {{\bf{p}},{\bf{q}}} \right) = \frac{{\exp \left( {\alpha  \cdot c\left( {{\bf{p}},{\bf{q}}} \right)} \right)}}{{\sum\limits_{{\bf{q}}' \in {\bf{Q}}} {\exp \left( {\alpha  \cdot c\left( {{\bf{p}},{\bf{q}}'} \right)} \right)} }},
%m(\mathbf{p},\mathbf{q}) = \frac{\exp(\alpha c(\mathbf{p},\mathbf{q}))}{\sum_{\mathbf{q}^{'} \in \mathbf{Q}}\exp(\alpha c(\mathbf{p},\mathbf{q}^{'}))},
\end{equation}
where $\mathbf{Q}$ is the set of 2D positions in $f^n$, and $\alpha$ is the temperature parameter. Note that soft-argmax converges to argmax as $\alpha$ increases, but an excessively high value of $\alpha$ can lead to unstable gradient flow during training. The optical flow field from the daytime to nighttime $F_{d\to n}$ is obtained as 
 \begin{equation}
\label{eqn: flow}
{F_{d \to n}}\left( {\bf{p}} \right) = \sum\limits_{{\bf{q}} \in {\bf{Q}}} {c'\left( {{\bf{p}},{\bf{q}}} \right) \cdot {\bf{q}}}.
\end{equation}
The optical flow field from the nighttime to daytime $F_{n\to d}$ is obtained in a similar manner by switching ${\bf{p}}$ and ${\bf{q}}$ in Eqs. (\ref{eqn: matching score})-(\ref{eqn: flow}). Finally, the semantic segmentation map warped from nighttime to daytime, denoted as $P_{n\to d}$, is obtained using $F_{d\to n}$ and $P_n$ by backward warping. Similarly, the semantic segmentation map warped from daytime to nighttime, denoted as $P_{d\to n}$, is obtained using $F_{n\to d}$ and $P_d$. Fig~\ref{fig:flow} shows some examples of $P_{n\to d}$ and $P_{d\to n}$. Although these warped predictions are imperfect, $P_{n\to d}$ ($P_{d\to n}$) is expected to be close to $P_{d}$ ($P_{n}$). Therefore, we can use $P_{n\to d}$ and $P_{d\to n}$ for the nighttime semantic segmentation network training.

\begin{figure}[t]
  \centering
  \includegraphics[width=0.95\linewidth]{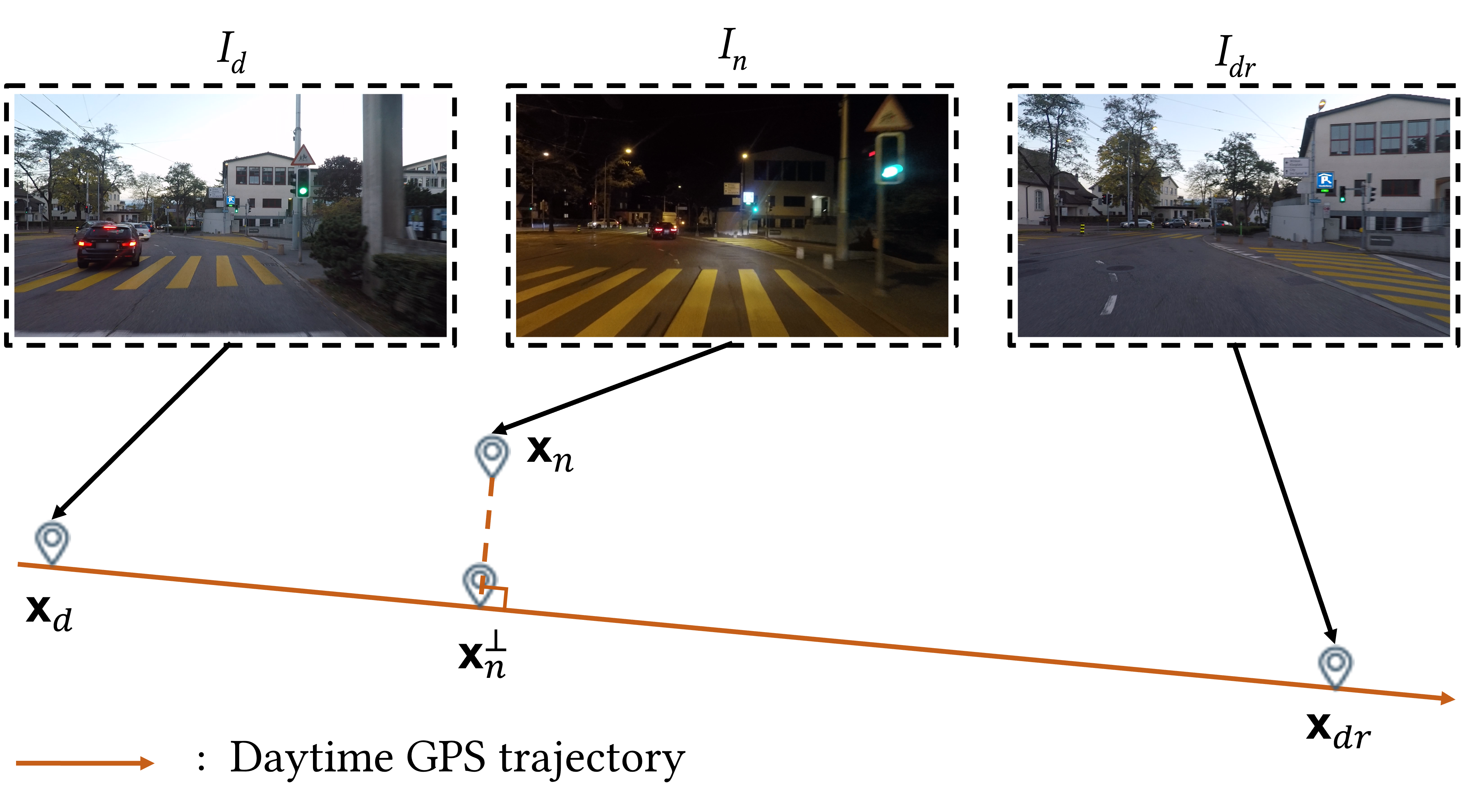}
  \caption{Illustration of the nighttime, daytime, and daytime reference images with their corresponding GPS positions. }
  \label{fig:gps_data}
\end{figure}

\subsubsection{Pseudo-supervision using intra-domain matching}
Due to the suboptimal performance of the relighting network, dense correspondence matching between daytime and nighttime is still challenging. Observing that most existing semantic image segmentation datasets~\cite{Cordts2016Cityscapes,Geiger2013kitti,sakaridis2020map-guided} provide video frames, we propose to use another daytime reference image, denoted as $I_{dr}$, for generating an additional pseudo-label. In the Dark Zurich dataset, $I_n$ is the nearest nighttime image of $I_d$, but the neighboring frames of $I_d$ along the forward and backward directions, denoted as $I_d^+$ and $I_d^-$, are also available. From the GPS positions of these images, we can determine $I_{dr}$ as either $I_d^+$ or $I_d^-$. 

Specifically, let ${{\bf{x}}_d}$, ${{\bf{x}}_d^+}$, ${{\bf{x}}_d^-}$, and ${{\bf{x}}_n}$ denote the GPS positions of $I_d$, $I_d^+$, $I_d^-$, and $I_n$, respectively. Here, each GPS position is given as a 2D vector containing the latitude and longitude. Then, $I_{dr}$ is determined as follows:
\begin{equation}
\label{eqn: cosine}
{I_{dr}} = \left\{ \begin{array}{l}
I_d^ + ,\,\,{\rm{if}}\,CS\left( {{{\bf{x}}_d} - {{\bf{x}}_n},{{\bf{x}}_d^ + } - {\bf{x}}_n} \right)\\ \,\,\,\,\,\,\,\,\,\,\,\,\,\, < CS\left( {{{\bf{x}}_d} - {{\bf{x}}_n},{{\bf{x}}_d^ - } - {\bf{x}}_n} \right),\\
I_d^ - ,\,\,{\rm{otherwise}},
\end{array} \right.
\end{equation}
where $CS$ measures the cosine similarity. In other words, as illustrated in Fig.~\ref{fig:gps_data}, if $I_n$ is located along the forward direction of $I_d$, $I_d^+$ is chosen as $I_{dr}$. Otherwise, $I_d^-$ is chosen as $I_{dr}$.

Given $I_d$ and $I_{dr}$, we obtain the optical flow field from the daytime to daytime reference, denoted as $F_{d\to dr}$, by following the same procedure of Eqs. (\ref{eqn: matching score})-(\ref{eqn: flow}) but with the features ${f^d}$ and ${f^{dr}}$, where ${f^{dr}} = \left\{ {f_l^{dr},f_g^{dr}} \right\}$ is the feature extracted from $I_{dr}$. The optical flow field from the daytime reference to daytime $F_{dr\to d}$ is obtained similarly. For the generation of the pseudo-label for the nighttime semantic segmentation network training, another pair of the optical flow fields are obtained as $F_{d \to n}^{'} = \lambda {F_{d \to dr}}$ and $F_{n \to d}^{'} = \lambda {F_{dr \to d}}$. The scale factor $\lambda$ is chosen as
\begin{equation}
\label{eqn: interpolation}
\lambda  = \frac{{HD\left( {{{\bf{x}}_d},{\bf{x}}_n^ \bot } \right)}}{{HD\left( {{{\bf{x}}_d},{{\bf{x}}_{dr}}} \right)}},
\end{equation}
where $HD$ measures the Haversine distance of two positions~\cite{inman1849haver}, and ${\bf{x}}_n^ \bot$ represents the position projected onto the line joining ${{\bf{x}}_d}$ and ${{\bf{x}}_{dr}}$, as illustrated in Fig.~\ref{fig:gps_data}. Finally, the semantic segmentation map warped from nighttime to daytime, denoted as $P^{'}_{n \to d}$, is obtained using $F^{'}_{d\to n}$ and $P_n$ by backward warping. Similarly, the semantic segmentation map warped from daytime to nighttime, denoted as $P^{'}_{d\to n}$, is obtained using $F^{'}_{n\to d}$ and $P_d$. Fig~\ref{fig:flow} shows some examples of $P^{'}_{n\to d}$ and $P^{'}_{d\to n}$. We now have four warped segmentation maps, i.e., $P_{n\to d}$, $P_{d\to n}$, $P^{'}_{n\to d}$, and $P^{'}_{d\to n}$, which are used for the nighttime semantic segmentation network training.

\subsubsection{Confidence map}

The first pair of the warped predictions, i.e., $P_{n\to d}$ and $P_{d\to n}$, can be inaccurate due to imperfect relighting and flow estimation. The second pair of the warped predictions, i.e., $P^{'}_{n\to d}$ and $P^{'}_{d\to n}$, can also be inaccurate because ${{\bf{x}}_n}$ is generally not lying on the line joining ${{\bf{x}}_d}$ and ${{\bf{x}}_{dr}}$, and thus the simple scaling by $\lambda$ can lead to imprecise flow fields. Moreover, GPS positions are not always precise. We thus define a 2D confidence map such that only consistent predictions are used for pseudo-supervision. Specifically, the confidence map for the nighttime to daytime warping, denoted as ${M_{n \to d}}$, is defined as follows:
\begin{equation}
\label{eqn: confidence_set}
{\bf{I(p)}}\!=\!\left\{{\bf{i}}|\arg\!\max \left( {{P_{n \to d}}\left( {\bf{p}} \right)} \right) \!=\!\arg\!\max \left( {{P^{'}_{n \to d}}\left( {{\bf{p}} \!+\! {\bf{i}}} \right)} \right) \right\},
\end{equation}
\begin{equation}
\label{eqn: confidence}
{M_{n \to d}}\left( {\bf{p}} \right) = \left\{ \begin{array}{l}
1,\;\;{\rm{if }}\;\;\exists {\bf{I(p)}} \in \Omega ,\\
0,\;\;{\rm{otherwise}},
\end{array} \right.
\end{equation}
where $\Omega$ is a set of positions in the 3$\times$3 kernel. ${{P_{n \to d}}\left( {\bf{p}} \right)}$ extracts the $C$-dimensional vector at ${\bf{p}}$, where $C$ is the number of semantic classes. The confidence map for the daytime to nighttime warping, denoted as ${M_{d \to n}}$, can be defined in a similar manner. These binary confidence maps are used when training the nighttime semantic network.

\subsection{Objective Functions}
We use five loss terms for GPS-GLASS: light loss~$L_{light}$, semantic segmentation loss~$L_{seg}$, adversarial loss~$L_{adv}$, discriminator loss~$L_{dis}$, and warping loss. Because we use the same loss functions defined in DANNet for the first four terms~\cite{wu2021dannet}, we only detail the warping loss in this subsection. 

We now have $P_{n\to d}$ and $P_{d\to n}$ and their confidence maps ${M_{n \to d}}$ and ${M_{d \to n}}$, which can be used to supervise the training of the nighttime semantic segmentation network. Note that $P^{'}_{n\to d}$ and $P^{'}_{d\to n}$ are integrated to $P_{n\to d}$ and $P_{d\to n}$ since only consistent predictions are used by the confidence maps. First, we use $P_{d\to n}$ for the pseudo-supervision of $P_{n}$. Specifically, the first warping loss term ${L_{d \to n}}$ is defined as follows:
\begin{equation}
\label{eqn: H}
{H({P_{d \!\to\! n}(\bf{q})},{P_{n}(\bf{q})})}\!=\!\sum\limits_{k \in \mathbb{C}}E_{o}{\left( { {P_{d \!\to\! n}}\left( {{\bf{q}};k} \right)} \right)\!\log {P_n}\left( {{\bf{q}};k} \right)},
\end{equation}
\begin{equation}
\label{eqn: L_dn}
{L_{d \to n}} \!=\!  - \frac{1}{{{N_p} \cdot C}}\!\sum\limits_{{\bf{q}} \in {{\bf{Q}}^{-}}} \!\!{{M_{d \to n}}\!\left( {\bf{q}} \right)}{H({P_{d \!\to\! n}(\bf{q})},{P_{n}(\bf{q})})} ,
\end{equation}
where $H$ measures the cross entropy, $E_{o}$ denotes the one-hot encoding~\cite{wu2021dannet}, $\mathbb{C}$ is a set of all semantic segmentation classes, $N_p$ is the number of pixels. ${{P_n}\left( {{\bf{q}};k} \right)}$ represents the probability of the $k$-th object class at the position ${\bf{q}}$ of ${P_n}$. Note that the cross-entropy loss is measured only for the reliable prediction with ${{M_{d \to n}}\left( {\bf{q}} \right)}=1$. Here, we define a set of ignore indexes, $\tilde {\bf{Q}}$, as follows:
\begin{equation}
\label{eqn: ignore}
\tilde {\bf{Q}}\!=\!\left\{ {{\bf{q}}\left| \begin{array}{l}
\!\arg\!\max \left( {{P_n}\left( {\bf{q}} \right)} \right)\!\in\!{C_{dyn}},\\
\!\arg\!\max \left( {{P_n}\left( {\bf{q}} \right)} \right)\!\ne\!\arg\!\max \left( {{P_{d \to n}}\left( {\bf{q}} \right)} \right)
\end{array} \right.}  \right\},
\end{equation}
where ${\mathbb{C}}_{dyn}$ is a set of dynamic semantic classes, including cars, people, etc. Then, ${{\bf{Q}}^{-}}$ in Eq. (\ref{eqn: ignore}) is defined as ${{\bf{Q}}^ - } = {\bf{Q}} \cap {\tilde {\bf{Q}}^c}$. We found this special handling is necessary to prevent undesirable pseudo-supervision of dynamic object classes. 

The second warping loss $L_{n\to d}$ is defined as follows:
\begin{equation}
\label{eqn: L_nd}
{L_{n \to d}} \!=\!  - \frac{1}{{{N_p} \cdot C}}\!\sum\limits_{{\bf{p}} \in {{\bf{P}}^{-}}} \!\!{{M_{n \to d}}\!\left( {\bf{p}} \right)}{H({P_{d}(\bf{p})},{P_{n \!\to\! d}(\bf{p})})} ,
\end{equation}
where ${{\bf{P}}^{-}}$ is defined in a similar manner as ${{\bf{Q}}^{-}}$. In other words, $P_{d}$ is used as the pseudo-supervision of $P_{n \to d}$ for the nighttime segmentation network training. 

The objective functions for the target daytime and nighttime domains, $L_{T_d}$ and $L_{T_n}$, and the source domain $L_{S}$ are defined as:
\begin{equation}
\label{eqn: L_day}
L_{T_d}=\mu_{1}L_{light}+\mu_{2}L_{adv},
\end{equation}
\begin{equation}
\label{eqn: L_night}
L_{T_n}=\mu_{1}L_{light}+L_{n \to d}+L_{d \to n}+\mu_{2}L_{adv},
\end{equation}
\begin{equation}
\label{eqn: L_source}
L_{S}=\mu_{1}L_{light}+\mu_{3}L_{seg}+\mu_{4}L_{dis},
\end{equation}
where $\mu_{1}$, $\mu_{2}$, $\mu_{3}$, and $\mu_{4}$ are empirically chosen as 0.01, 0.01, 1, and 1, respectively, following the baseline method DANNet~\cite{wu2021dannet}. In every training iteration of GPS-GLASS, we sequentially optimize $L_{T_d}$, $L_{T_n}$, and $L_{S}$ for daytime, nighttime, and source domains, respectively.

\begin{figure}[t]
  \centering
  \includegraphics[width=\linewidth]{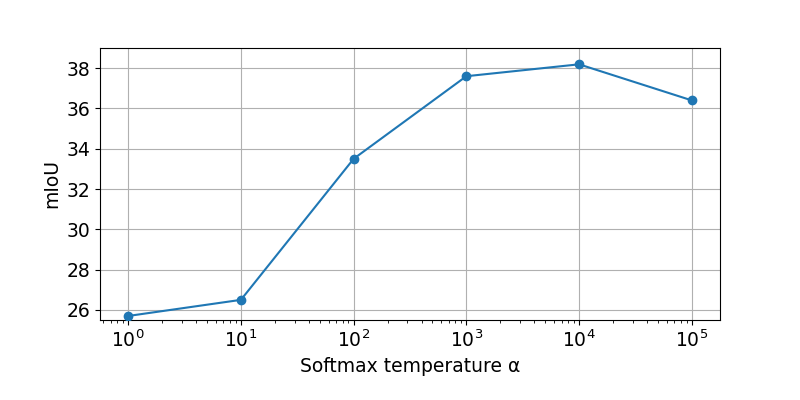}
  \caption{Performance evaluation with different softmax temperature values on Dark Zurich-val.}
  \label{figure:temperature}
\end{figure}

\begin{table*}[t]
\centering
\renewcommand{\tabcolsep}{.7mm}
\caption{Performance comparison on ACDC-night. The best and second-best results are boldfaced and underlined, respectively.}
\label{table:ACDC_test}
\begin{tabular}{lllllllllllllllllllll}
\toprule
Method&\rotatebox[origin=c]{90}{road}&\rotatebox[origin=c]{90}{sidewalk}.&\rotatebox[origin=c]{90}{building}&\rotatebox[origin=c]{90}{wall}&\rotatebox[origin=c]{90}{fence}&\rotatebox[origin=c]{90}{pole}&\rotatebox[origin=c]{90}{light}&\rotatebox[origin=c]{90}{sign}&\rotatebox[origin=c]{90}{vegetation}.&\rotatebox[origin=c]{90}{terrain}&\rotatebox[origin=c]{90}{sky}&\rotatebox[origin=c]{90}{person}&\rotatebox[origin=c]{90}{rider}&\rotatebox[origin=c]{90}{car}&\rotatebox[origin=c]{90}{truck}&\rotatebox[origin=c]{90}{bus}&\rotatebox[origin=c]{90}{train}&\rotatebox[origin=c]{90}{motocycle}&\rotatebox[origin=c]{90}{bicycle}&\bf mIoU\\
\midrule
RefineNet& 66.8& 24.0& 50.3& 16.9& 11.6& 26.4& 34.2& 25.5& 44.2& 21.6& 0.1& 40.8& 24.8& 57.4& 6.8& 37.3& 20.5& 24.0& 19.1& 29.1\\
DeepLabV2& 77.0& 22.9& 56.3& 13.5& 9.2& 23.8& 22.9& 25.6& 41.4& 16.1& 2.9& 44.2& 17.5& 64.1& 11.9& 34.5& 42.4& 22.7& 22.7& 30.1\\
PSPNet& 75.5& 16.3& 47.3& 14.5& 10.4& 23.2& 29.0& 22.8& 40.5& 10.8& 12.0& 39.2& 15.3& 44.3& 2.6& 23.0& 37.5& 13.8& 27.9& 26.6\\
\midrule
DMAda& 74.7& 29.5& 49.4& 17.1& 12.6& 31.0& 38.2& 30.0& 48.0& 22.8& 0.2& 47.0& 25.4& 63.8& 12.8& 46.1& 23.1& 24.7& 24.6& 32.7\\
GCMA& 78.6 &45.9 &58.5 &17.7 &18.6 &\underline{37.5} &\bf43.6 &\bf43.5 &58.7 &39.2 &22.5 &\bf57.9 &29.9 &\bf72.1 &21.5 &\underline{56.3} &41.8 &\underline{35.7} &35.4 &42.9\\
MGCDA& 74.5 &52.5 &69.4 &7.7 &10.8 &\bf38.4 &\underline{40.2} &\underline{43.3} &61.5 &36.3 &37.6 &\underline{55.3} &25.6 &\underline{71.2} &10.9 &46.4 &32.6 &27.3 &33.8 &40.8\\
DANNet& 90.7& \underline{61.2}& 75.6& 35.9& 28.8& 26.6& 31.4& 30.6& \bf70.8& \underline{39.4}& 78.7& 49.9& 28.8& 65.9& 24.7& 44.1& \bf61.1& 25.9& 34.5& 47.6\\
DANIA&\underline{91.0} &60.9 &\bf77.7 &\underline{40.3} &\bf30.7 &34.3 &37.9 &34.5 &\underline{70.0} &37.2 &\underline{79.6} &45.7 &\bf32.6 &66.4 &11.1 &37.0 &\underline{60.7} &32.6 &\bf37.9 &48.3\\
CCDistill & 90.0& 60.7& 75.6& \bf{42.0}& 28.3& 27.5& 29.2& 32.2& 67.7& 36.0& 77.4& 46.7& 24.2& 69.7& \bf{48.2}& 45.4& 53.9& \bf{40.5}& 36.0& \underline{49.0}\\
\midrule
\bf GPS-GLASS& \bf91.8& \bf65.0& \underline{76.4}& 38.1& \underline{30.0}& 35.8& 38.5& 37.6& 69.2& \bf41.4& \bf79.8& 45.8& \underline{31.2}& 69.6& \underline{38.0}& \bf59.9& 45.7& 24.9& \underline{37.2}& \bf50.3\\
\bottomrule
\end{tabular}
\end{table*}

\section{Experimental Results}
\subsection{Datasets}
The \textbf{Cityscapes dataset~\cite{Cordts2016Cityscapes}} includes 5,000 images taken in street scenes with pixel-level annotations for a total of 19 categories.  In total, there are 2,975 images for training, 500 images for validation, and 1,525 images for testing. We used the training set as the labeled source domain $S$ in the GPS-GLASS training stage. %A comparison of catastrophic forgetting in the daytime was performed on the validation set.

\noindent The \textbf{Dark Zurich dataset~\cite{sakaridis2019guided}} includes 2,416 nighttime images, 2,920 twilight images and 3,041 daytime images for training, which are all unlabeled with the size of 1,920$\times$1,080. The images across different domains are coarsely paired according to the GPS distance-based nearest neighbor assignment. Consequently, most of these images share many image contents that are valuable for domain adaptation in semantic segmentation. Following the previous works~\cite{wu2021dannet,xu2021cdada}, we only used 2,416 day-night image pairs in the training stage of GPS-GLASS as the unlabeled target domains, $T_d$ and $T_n$. For the quantitative performance evaluation, the Dark Zurich dataset provides 50 finely annotated nighttime images, which are also used for our ablation study. 

\noindent The \textbf{ACDC-night dataset~\cite{sakaridis2021acdc}} is an extended version of the Dark Zurich dataset, including 1006 nighttime images (400, 106, and 500 images for training, validation, and test). The dataset also provides finely annotated nighttime images as the Dark Zurich dataset. The performance evaluation on the ACDC-night dataset was conducted on a test set using an online evaluation website~\cite{acdcweb}.

\noindent The \textbf{NightCity+ dataset~\cite{deng2022nightlab}} is an extended version of the NightCity dataset~\cite{tan2021night} that re-annotates incorrectly labeled regions of the validation set. The NightCity+ dataset provides 2998 and 1299 images for training and validation, respectively, which were taken from nighttime street scenes in various cities. We used the NightCity+ validation dataset only for the performance evaluation.

\begin{figure}[t]
  \centering
  \includegraphics[width=\linewidth]{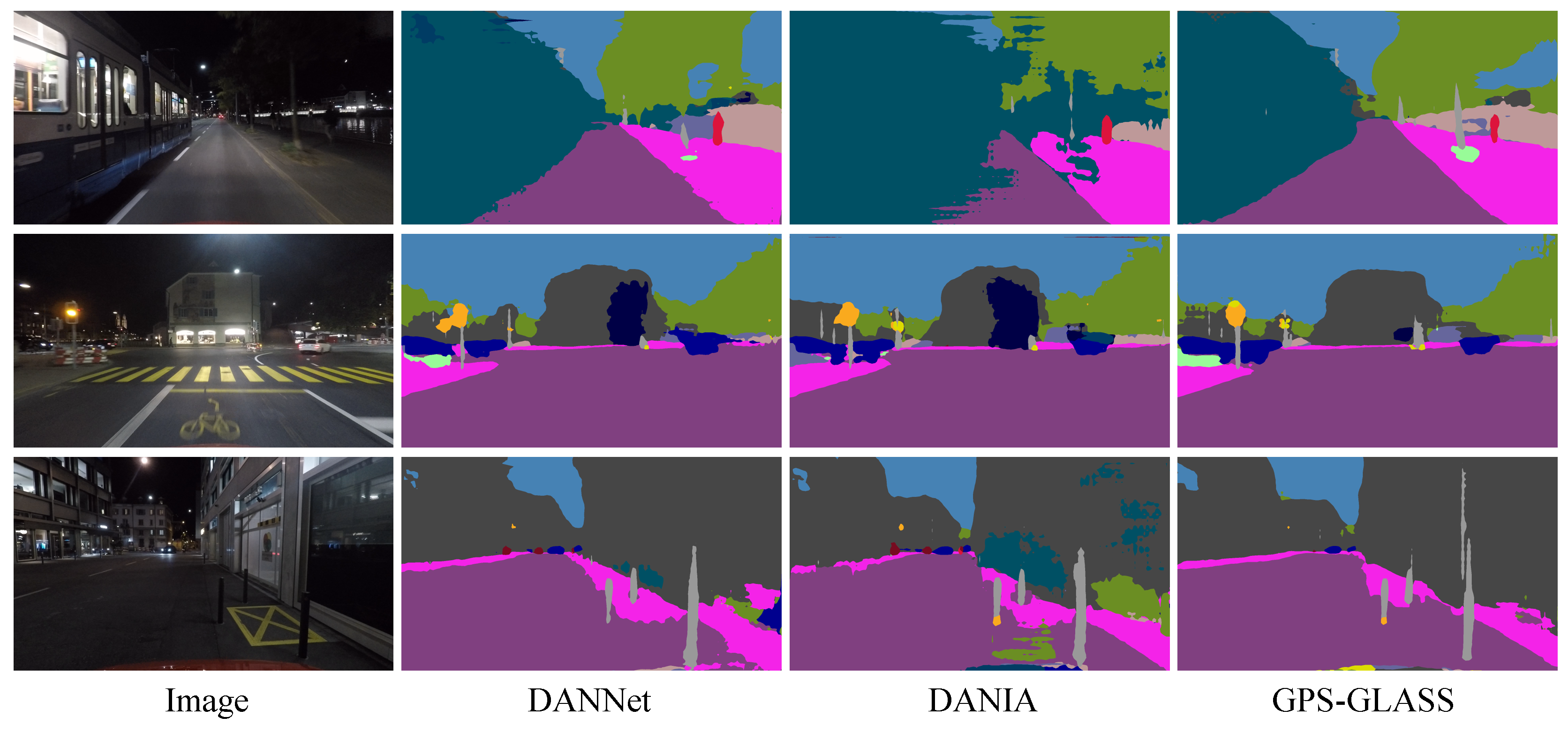}
  \caption{Visual comparison of our GPS-GLASS with other state-of-the-art methods on ACDC-night.}
  \label{figure:acdc_result}
\end{figure}

\subsection{Implementation Details}
We implemented GPS-GLASS using PyTorch. The training was performed with a single Nvidia Titan RTX GPU. Following~\cite{chen2017deeplab}, we trained our network using the stochastic gradient descent optimizer with a momentum of 0.9 and a weight decay of $5 \times 10^{-4}$. We used Adam optimizer~\cite{kingma2014adam} for training the discriminators with $\beta$ of 0.9 and 0.99. The initial learning rate of the generator and discriminators was set to $2.5 \times 10^{-4}$ and then reduced to the power of 0.9 using the poly learning rate policy. For data augmentation, random cropping of the size $512\times512$ was applied with a scale factor between 0.5 and 1.0 for the Cityscapes dataset, and random cropping of the size $960\times960$ was applied with a scale factor between 0.9 and 1.1 for the Dark Zurich dataset. In addition, we applied random horizontal flips for training. We used PSPNet~\cite{zhao2017pyramid} as the segmentation network model, which has shown state-of-the-art performance in nighttime semantic segmentation. We pre-trained PSPNet on the Cityscapes dataset for 150K iterations using $L_{seg}$. Then, we set the batch size to 2 and trained the model for 100K iterations.

\begin{table}[t!]
  \centering
  \renewcommand{\arraystretch}{.9}
  \caption{Performance comparison on Dark Zurich-val and NightCity+.}
  \begin{tabular}{lcc}
  \toprule\
        \multirow{2}{*}{Method}&\multicolumn{2}{c}{mIoU}\\ \cmidrule{2-3}
        &Dark Zurich-val&NightCity+\\
        \midrule
        PSPNet&12.28&19.04\\
        \midrule
        GCMA&26.65&-\\
        MGCDA&26.10&-\\
        DANNet&36.76&29.93\\
        DANIA&  38.14&28.92\\
        \midrule
        \bf GPS-GLASS&\bf 38.19&\bf31.81\\
        \bottomrule
    \end{tabular}
    \label{table:other_dataset}
\end{table}

We found that careful selection of the temperature value $\alpha$ in Eq. (\ref{eqn: softmax}) is important for correspondence matching. The performance evaluation on Dark Zurich-val for GPS-GLASSes trained with different $\alpha$ values is shown in Fig.~\ref{figure:temperature}. From this grid search of $\alpha$, we chose $\alpha=10^{4}$ in our experiments. 

\subsection{Performance Comparisons}

\begin{table*}[htb]
\centering
\renewcommand{\tabcolsep}{0.65mm}
\renewcommand{\arraystretch}{.9} 
\caption{Ablation study on several model variants of our method on Dark Zurich-val.}
\label{table:ablation}
\begin{tabular}{l|ccc|cc|cc|cc|c|cc}
\hline
                                    &avg & max&confidence&inter & intra   & $L_{n \to d}$    &$L_{d \to n}$     & local feature     & global feature    &static loss & mIoU    &Gain \\
\hline       
w/o pseudo-supervision               &&&                   &     &                 &               &               &                   &                   &            & 24.68\\
\hline
DANNet

&&&                &       &                 &               &               &                   &                   &\checkmark  &36.76   &+12.08\\
\hline 
\multirow{2}{*}{inter-intra mixing }                         & \checkmark      &    &   & \checkmark    & \checkmark        & \checkmark        &         \checkmark  & \checkmark  & \checkmark &       & 35.92  &+11.24\\
                          &   &   \checkmark&     & \checkmark    & \checkmark        & \checkmark        & \checkmark       & \checkmark  & \checkmark  &          &35.70    &+11.02   \\
\hline
\multirow{2}{*}{matching domain }         &     &   &              & \checkmark     &             & \checkmark    & \checkmark    & \checkmark        & \checkmark        &            & 34.86  &+10.18\\
          &     &   &            &   & \checkmark        & \checkmark    & \checkmark    & \checkmark        & \checkmark        &            &36.21    &+11.53   \\
\hline
\multirow{2}{*}{w/o one warping loss}   &     &   &\checkmark       &\checkmark     &\checkmark     & \checkmark    &               & \checkmark        & \checkmark       &  &  33.90          &+9.22      \\
                     &     &   &\checkmark&\checkmark    &\checkmark      &               & \checkmark    & \checkmark        & \checkmark        &            &36.06    &+11.38    \\
\hline
\multirow{2}{*}{w/o global/local feature}&     &   &\checkmark  &\checkmark   &\checkmark       & \checkmark    & \checkmark    & \checkmark        &                   &            &37.45          &+12.77      \\
                     &     &   &\checkmark&\checkmark   &\checkmark        & \checkmark    & \checkmark    &                   & \checkmark        &                   &36.67            &+11.99      \\
\hline
\textbf{GPS-GLASS}    &     &   &\checkmark       &\checkmark       & \checkmark   & \checkmark    & \checkmark    & \checkmark        & \checkmark        &            & \bf{38.19}  &\bf{+13.51}  \\
\hline
\end{tabular}
\end{table*}

\subsubsection{Comparison on ACDC-night and Dark Zurich}
We compared GPS-GLASS with several state-of-the-art domain adaptation-based nighttime semantic segmentation methods, including DANNet~\cite{wu2021dannet}, DANIA~\cite{wu2021dania}, MGCDA~\cite{sakaridis2020map-guided}, GCMA~\cite{sakaridis2019guided}, DMAda~\cite{dai2018DMAda}, and CCDistill~\cite{gao2022ccdistill}. For the comparison with the other techniques, BDL, AdaptSegNet, ADVENT~\cite{tsai2018adaptseg,vu2019advent,li2019bidirectional} were also evaluated, where they were trained to adapt from Cityscapes to Dark Zurich. We report the mean intersection over union (mIoU) as the evaluation metric. For accurate performance comparison, we used ACDC-night, an extended version of Dark Zurich, which provides a large number of images with difficult object classes to be segmented. Table~\ref{table:ACDC_test} reports the mIoU results on ACDC-night. All of these compared methods used the common ResNet-101~\cite{he2016resnet101} as a backbone. We used DANNet and DANIA with PSPNet~\cite{zhao2017pyramid} for a fair comparison with our GPS-GLASS. 

GPS-GLASS achieved a 1.3\% performance improvement in terms of the mIoU over the second-best method, CCDistill. Note that GPS-GLASS does not increase the number of network parameters or processing time compared to DANNet because the same architecture of PSPNet is used in the inference stage. The performance improvements are significant in several categories, such as road, sidewalk, terrain, and sky, which are difficult to identify in nighttime scenes. Meanwhile, due to pixel-level aligned pseudo-supervision, improvements are also noticeable in small-scale classes such as poles, lights, and sign, compared to the baseline method, DANNet. Consistent results were obtained from Dark Zurich-val as shown in Table \ref{table:other_dataset}. These results indicate that our approach effectively performed the domain adaptation from the daytime to nighttime. Fig.~\ref{figure:acdc_result} shows several results for visual comparison. More frame-by-frame comparisons are provided in our project page.

\subsubsection{Generalization Ability for Nighttime}
To show the generalization ability of our proposed method, we tested our model trained on Dark Zurich to NightCity+. As shown in Table \ref{table:other_dataset}, GPS-GLASS achieved a 1.88\% performance improvement in terms of the mIoU over the second-best method, DANNet. This result demonstrates that the proposed GPS-GLASS trained on Dark Zurich generalizes well to another challenging nighttime dataset. %Visual comparisons are provided on our project page. 

\begin{figure}[t]
  \centering
  \includegraphics[width=\linewidth]{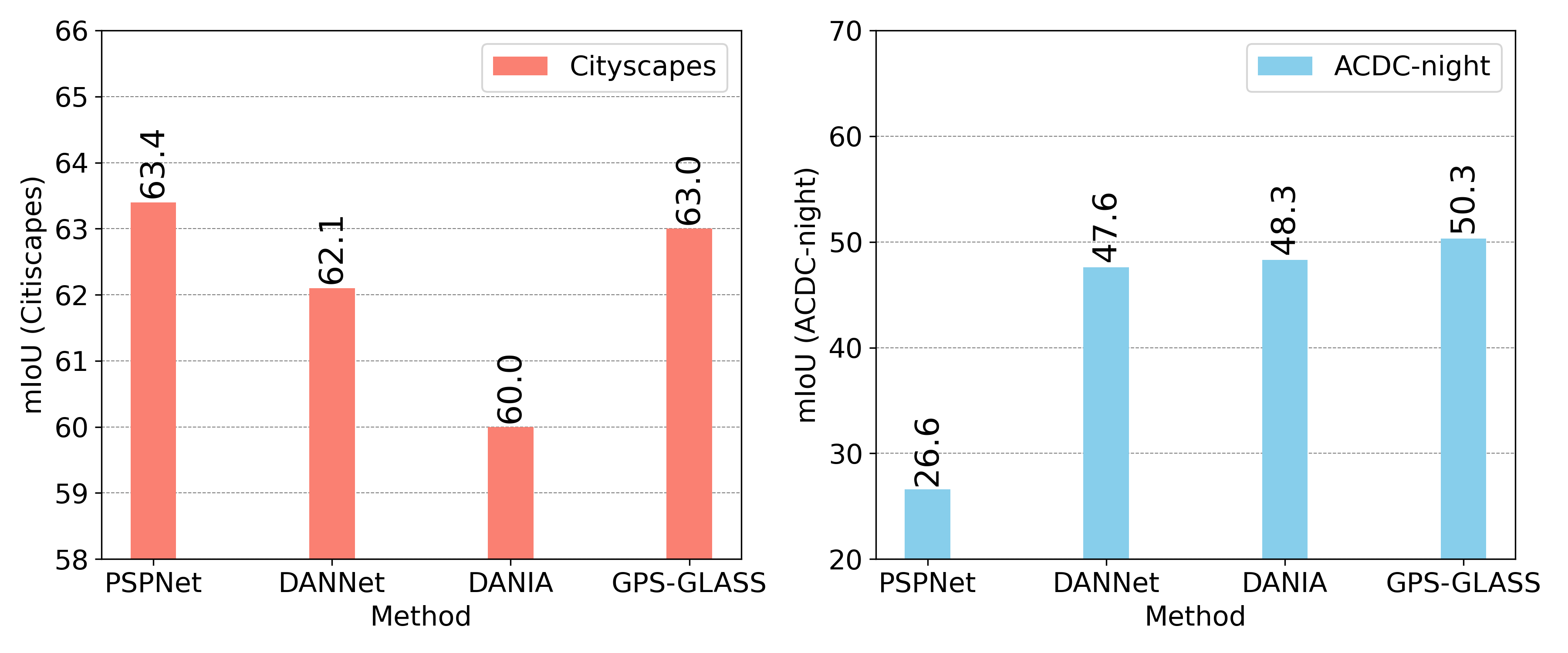}
  \caption{Performance comparison of daytime and nighttime semantic segmentation results for our GPS-GLASS and other state-of-the-art methods.}
  \label{figure:city_result}
\end{figure}

\subsubsection{Generalization Ability for Daytime}
One of the technical challenges of nighttime domain adaptation is the generalization ability for daytime. To this end, we compared the proposed GPS-GLASS with PSPNet~\cite{zhao2017pyramid} (trained on Cityscapes) as well as DANNet and DANIA (pre-trained on Cityscapes and then trained with Dark-Zurich via UDA). As shown in Fig.~\ref{figure:city_result}, DANNet and DANIA showed noticeable performance drops from PSPNet on Cityscapes. However, our proposed GPS-GLASS achieved state-of-the-art performance at nighttime at the sacrifice of a 0.4\% performance reduction at daytime. We consider that the correlation layer of GPS-GLASS enabled the segmentation network to extract domain-invariant features through correspondence matching, resulting in high performance across two domains.

%This is due to the behavior of the network forgetting previously trained knowledge when training on new domain. 

\subsection{Ablation Study}
In order to demonstrate the effectiveness of individual components of GPS-GLASS, several modified models of GPS-GLASS were trained, and the best performances in Dark Zurich-val are reported in Table~\ref{table:ablation}. GPS-GLASS without any pseudo-supervision serves as a naive baseline, which leads to the lowest mIoU of 24.68. Due to the static loss~\cite{wu2021dannet}, DANNet achieved a 12.08\% mIoU increase compared to the baseline. We applied other inter-intra pseudo-label mixing methods: taking the average of two pseudo-labels or taking the label with the higher probability for each pixel. For Dark Zurich-val, these two methods, denoted as avg and max in Table~\ref{table:ablation}, increased the mIoU by 11.24\% and 11.02\%, respectively, which are worse than the performance improvement obtained using the confidence map (13.51\%). Both warping loss terms, $L_{n\to d}$ and $L_{d\to n}$, were found to be essential compared to their single-use. In addition, because we obtained the integrated correlation volume by element-wise multiplication of the correlation volumes from the local and global features, we evaluated the performance obtained without using the local or global feature. The use of both features resulted in 0.74\% or 1.52\% higher mIoU compared to the single-use of the local or global feature, respectively. Last, because GPS-GLASS obtains pseudo-supervision from both intra-matching and inter-matching, we evaluated the performance without applying intra-matching or inter-matching and obtained 3.33\% or 1.98\% lower mIoU compared to GPS-GLASS, respectively. 

\section{Conclusions}
In this paper, we proposed GPS-GLASS, a novel training methodology for nighttime semantic segmentation based on unlabeled daytime-nighttime image pairs and their GPS data. GPS-GLASS obtains pixel-level aligned pseudo-supervision through bidirectional correspondence matching between the daytime and nighttime. To address the difficulty of correspondence matching between different domains, GPS-GLASS also acquires another pseudo-supervision through correspondence matching in the same daytime domain using the GPS data. The confidence map is used to exclude pseudo-supervision of less reliable predictions. Our GPS-GLASS does not increase the number of network parameters or inference time compared to the adopted baseline model. Experimental results on the ACDC-night, Dark Zurich-val, and NightCity+ datasets demonstrate the effectiveness of the proposed method.

\newpage

%%%%%%%%% REFERENCES
{\small
\bibliographystyle{ieee_fullname}
\bibliography{egbib}
}
\end{document}

\title{GPS-GLASS: Learning Nighttime Semantic Segmentation Using Daytime Video and GPS data\\  % **** Enter the paper title here
\textit{-Supplementary Material-}}
\author{Hongjae Lee, Changwoo Han, Jun-Sang Yoo, Seung-Won Jung\thanks{Corresponding author}\\
	Department of Electrical Engineering, Korea University\\
	{\tt\small \{jimmy9704, hcwoo329, junsang7777, swjung83\}@korea.ac.kr}
}

\maketitle
\thispagestyle{empty}
\appendix

%%%%%%%%% BODY TEXT - ENTER YOUR RESPONSE BELOW

\section{Introduction}
For a better understanding of GPS-GLASS, this supplementary material includes an in-depth analysis of the GPS data (Section~\ref{section:gps}) and correspondence matching (Section~\ref{section:correspondence}). Moreover, more visualizations (Section \ref{section:viz}) are provided. 

\section{Analysis of the GPS data}
\label{section:gps}
\begin{figure}[h!]
\centering
\includegraphics[width=0.95\linewidth]{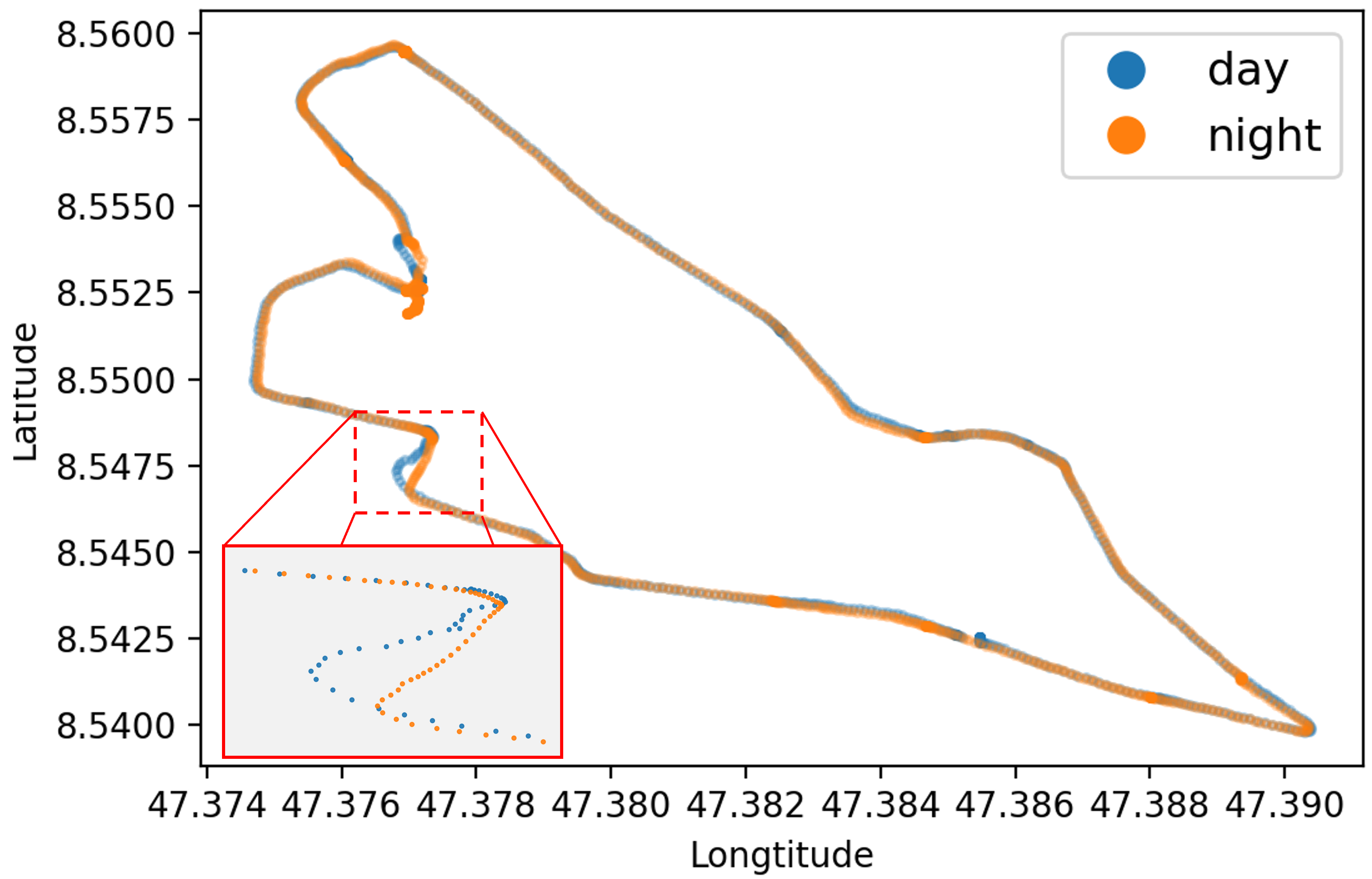}
\caption{Visualization of the GPS route of the daytime (blue dots) and nighttime (orange dots) images in Dark Zurich-train.}
\label{fig:gps_route}
\end{figure}

\begin{figure}[h!]
\centering
\includegraphics[width=0.9\linewidth]{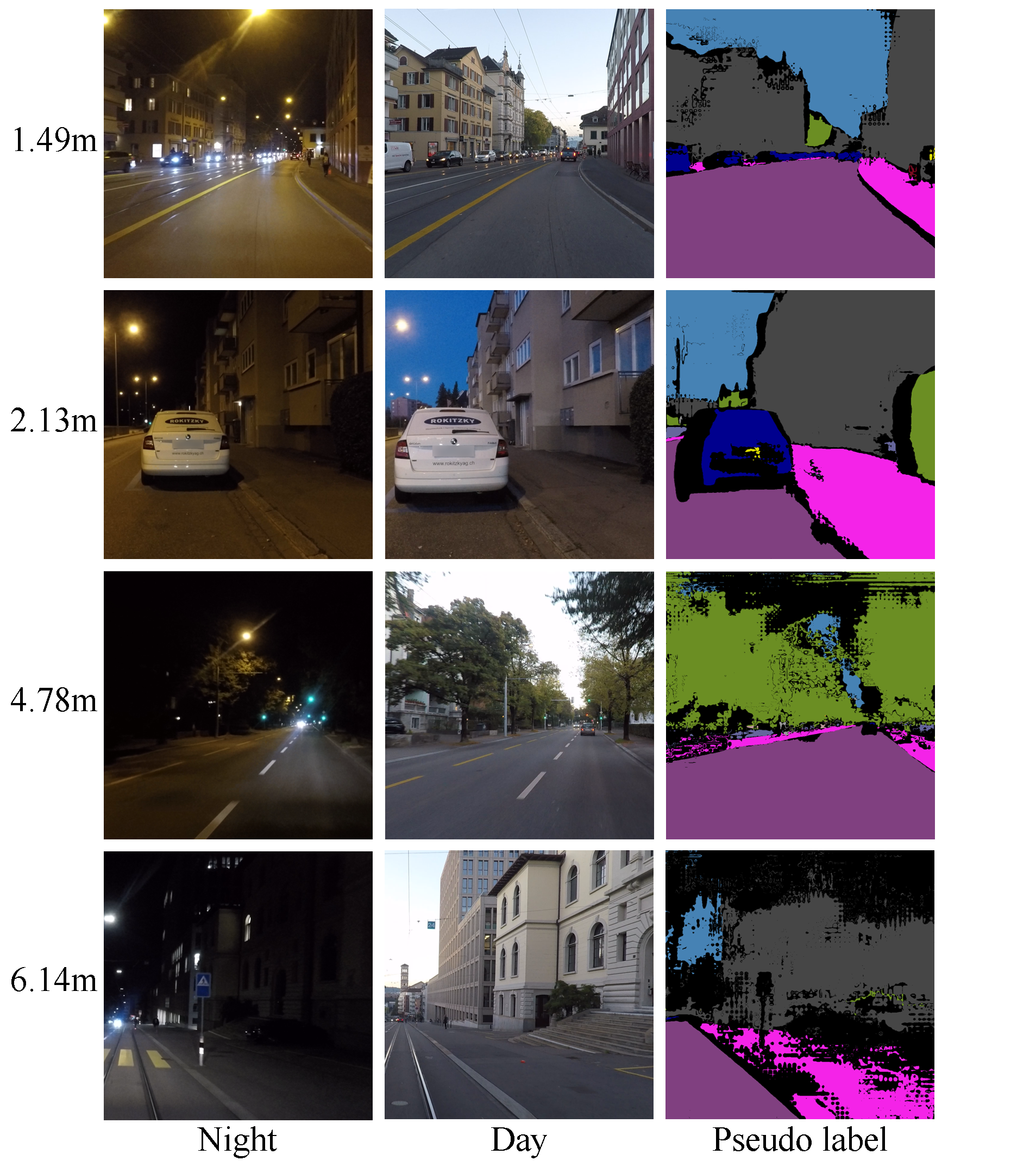}
\caption{Visualization of the pseudo-labels with confidence maps in Dark Zurich-train. The pixels with zero values in the confidence maps are marked as black in the pseudo-labels. The GPS distance was computed as the Haversine distance between daytime and nighttime GPS coordinates.}
\label{fig:confidence}
\end{figure}

\begin{figure}[h!]
\centering
\includegraphics[width=0.94\linewidth]{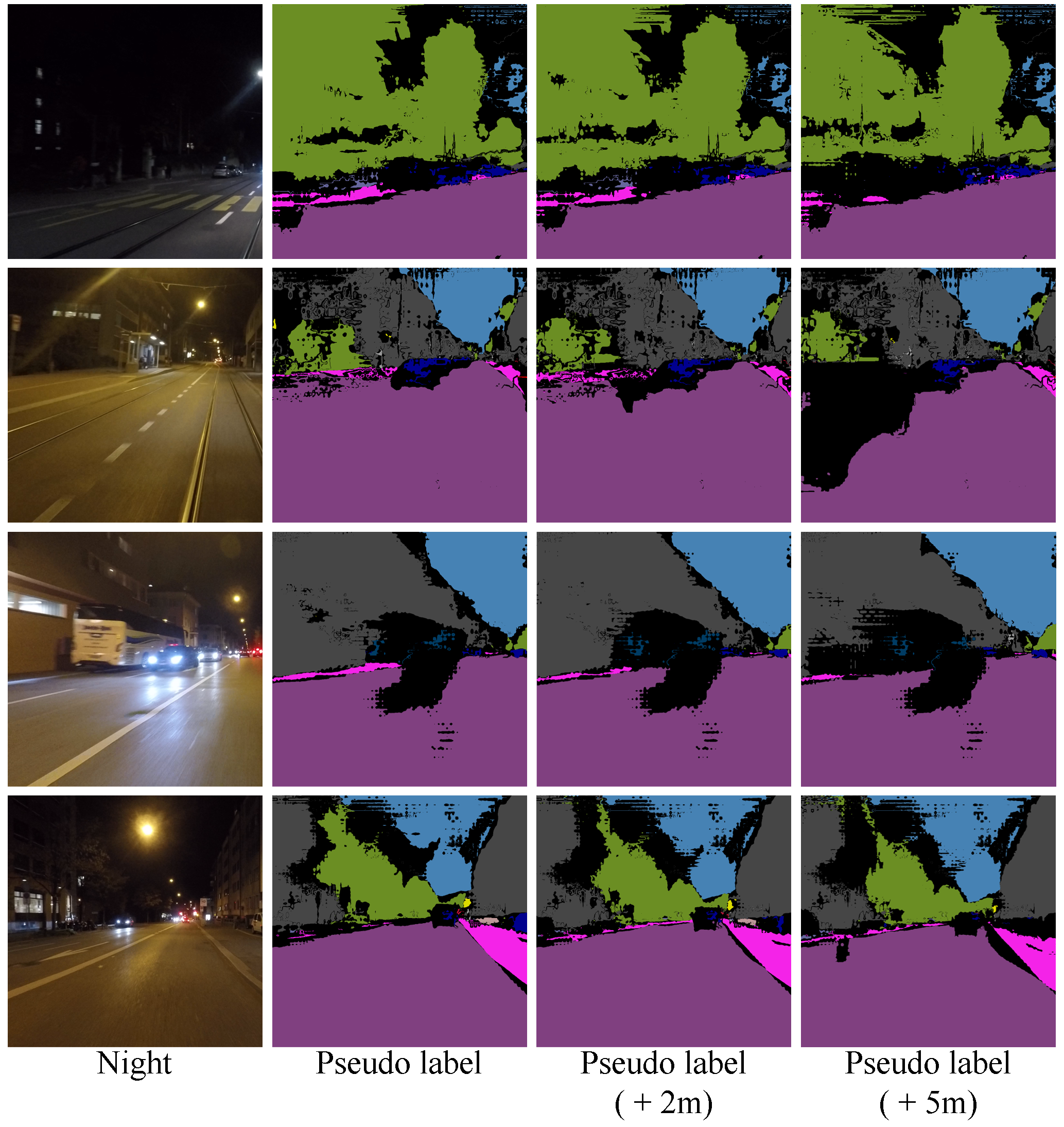}
\caption{Visualization of the pseudo-labels with confidence maps according to each noise in Dark Zurich-train.}
\label{fig:GPS_error}
\end{figure}

As shown in Fig.~\ref{fig:gps_route}, the paths of daytime and nighttime are very similar, providing reliable image pairs for training GPS-GLASS. However, several inaccurate samples still significantly differ between the daytime and nighttime, \textit{e.g.}, those in the red box in Fig.~\ref{fig:gps_route}.  Because GPS information is used only at the training stage, we proposed to use a confidence map to minimize the effect of any possible inaccurately estimated pseudo-labels during the training. Specifically, by Eq. (6), most pixels corresponding to inaccurate pseudo-labels have zero values in the confident maps, making the network not use such pixels for the training. Several examples of the pseudo-labels and confidence maps with different distances between nighttime and daytime images are shown in Fig.~\ref{fig:confidence}. Notice that the number of pixels with zero values in the confidence maps increases due to occlusions as the distance between daytime and nighttime images increases.

To test the robustness of the proposed method against GPS errors, we synthesized noise to the GPS data. Specifically, we changed the nighttime GPS to be further 2m and 5m away from the daytime GPS and applied the proposed method. As shown in Fig.~\ref{fig:GPS_error}, the number of pixels with zero values in the confidence maps increases as the GPS error increases. These results demonstrate that the confidence map can effectively eliminate inaccurate pixels from pseudo-labels.

\begin{figure*}[h!]
  \centering
  \includegraphics[width=0.8\linewidth]{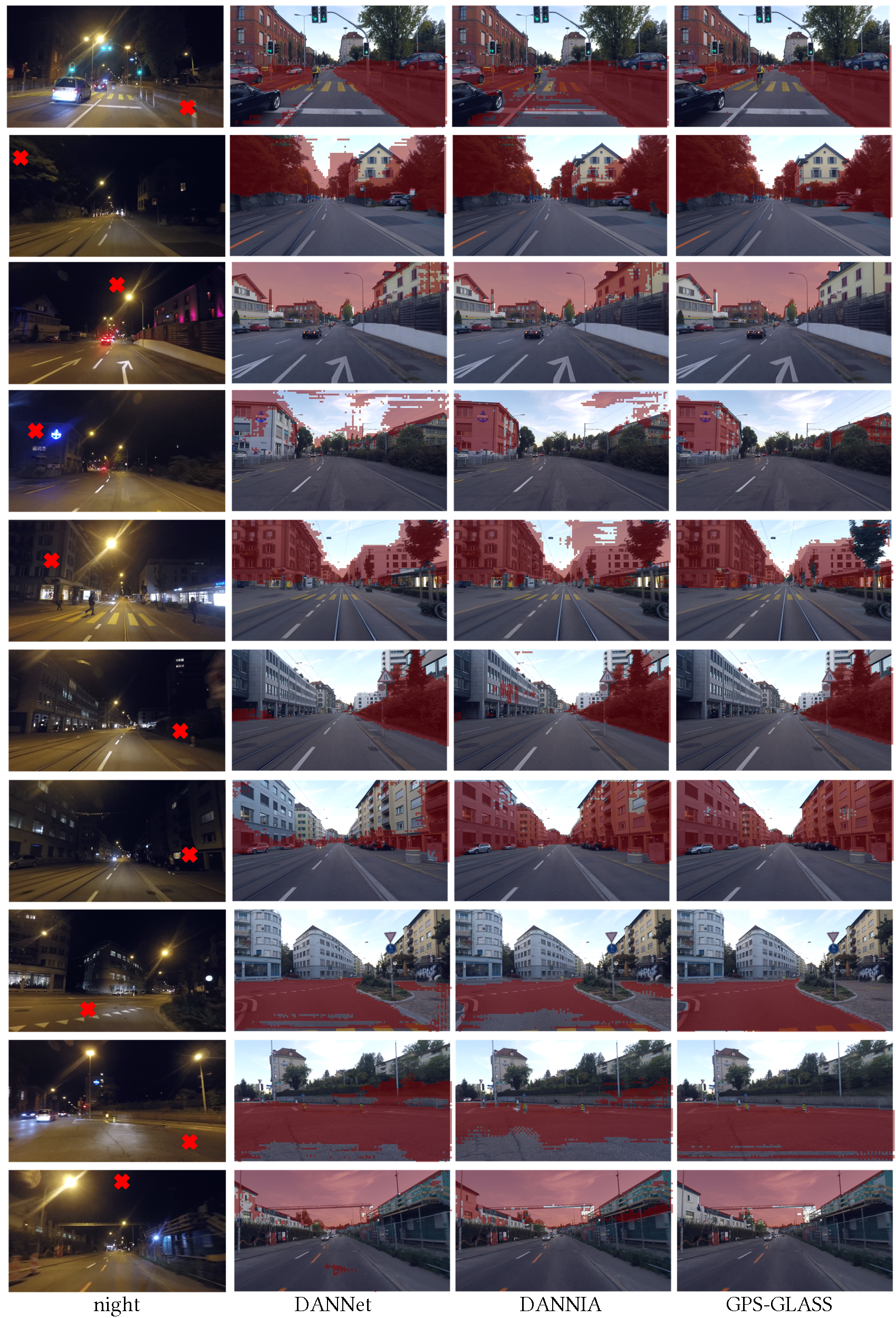}
  \caption{Comparisons of the correspondence matching results obtained by the proposed GPS-GLASS, DANNet, and DANIA. We computed the cosine similarity between the nighttime feature at the red cross position and the daytime features. The pixels in the daytime images with similarities above the threshold are colored in red.}
  \label{figure:correspondence}
\end{figure*}

\section{Analysis of Correspondence Matching}
\label{section:correspondence}

We conducted additional analysis for dense correspondence matching between nighttime and daytime images. To this end, we computed the cosine similarity between the feature of the nighttime image at the red cross position, as shown in Fig.~\ref{figure:correspondence}, and the features of the daytime image, where the features were obtained before the last layer of the segmentation network. The pixels with similarity score higher than the threshold are colored in red, where the threshold was set to 0.25 for all methods and scenes. As shown in Fig.~\ref{figure:correspondence}, GPS-GLASS assigned high similarities for the pixels with the semantic class. However, DANNet and DANIA resulted in high similarities for many pixels belonging to different semantic classes (e.g., building, sky, sidewalk, and road). These results indicate that the correlation layer drives the segmentation network to extract features with high similarities to the pixels with the same semantic classes between daytime and nighttime images.

\section{Visual Comparisons}
\label{section:viz}
\begin{figure*}[t]
  \centering
  \includegraphics[width=\textwidth]{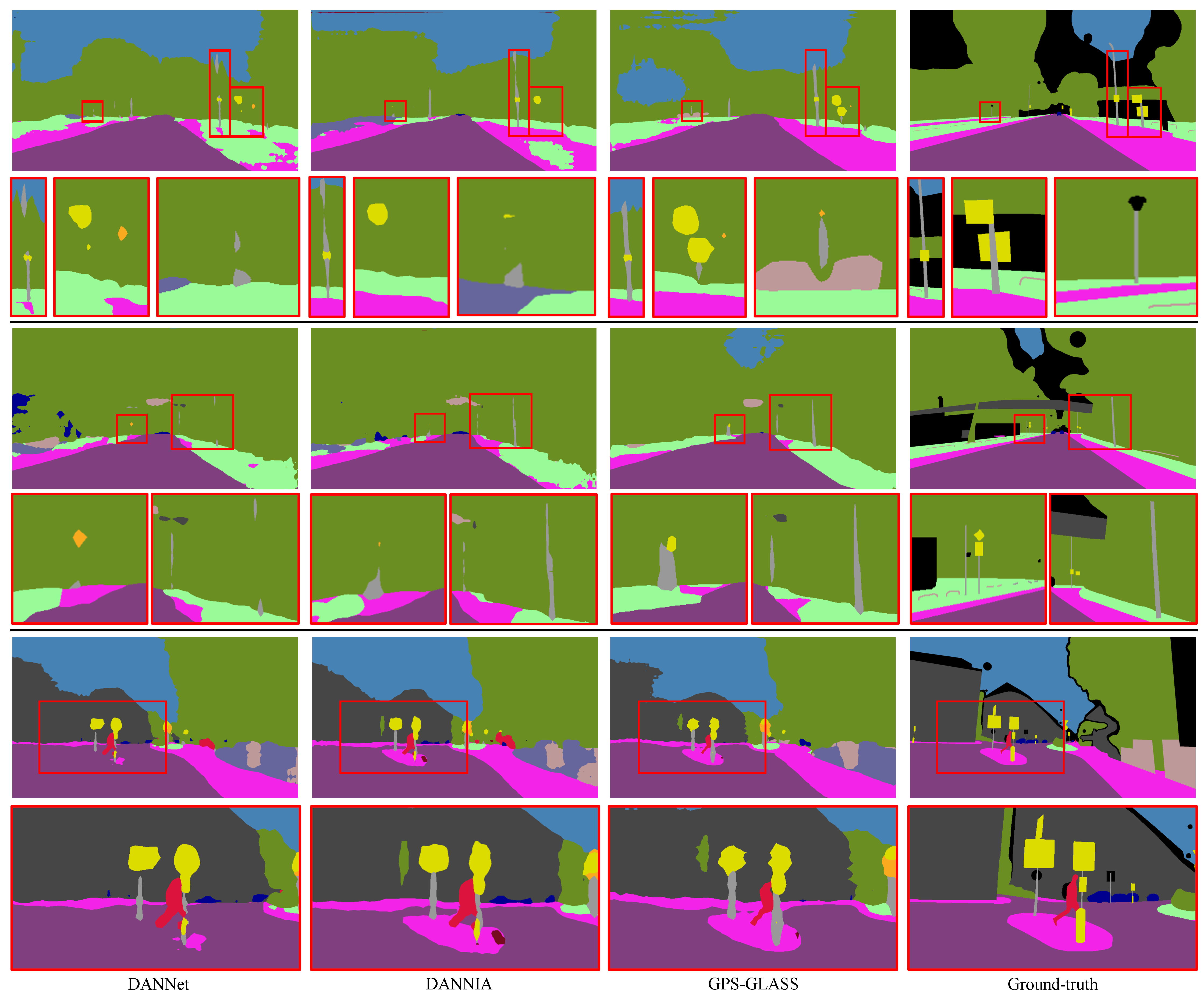}
  \caption{More results of our proposed GPS-GLASS, DANNet, and DANIA for visual comparison. The results in the red boxes are zoomed in for better visualization. Note that small object classes, such as pole and lights, are more precisely segmented by GPS-GLASS compared to DANNet and DANIA. In addition, object classes sharing similar features, such as road and sidewalk, are well separated by GPS-GLASS.}
  \label{figure:small_scale}
\end{figure*}

\begin{figure*}[h!]
\centering
\includegraphics[width=\textwidth]{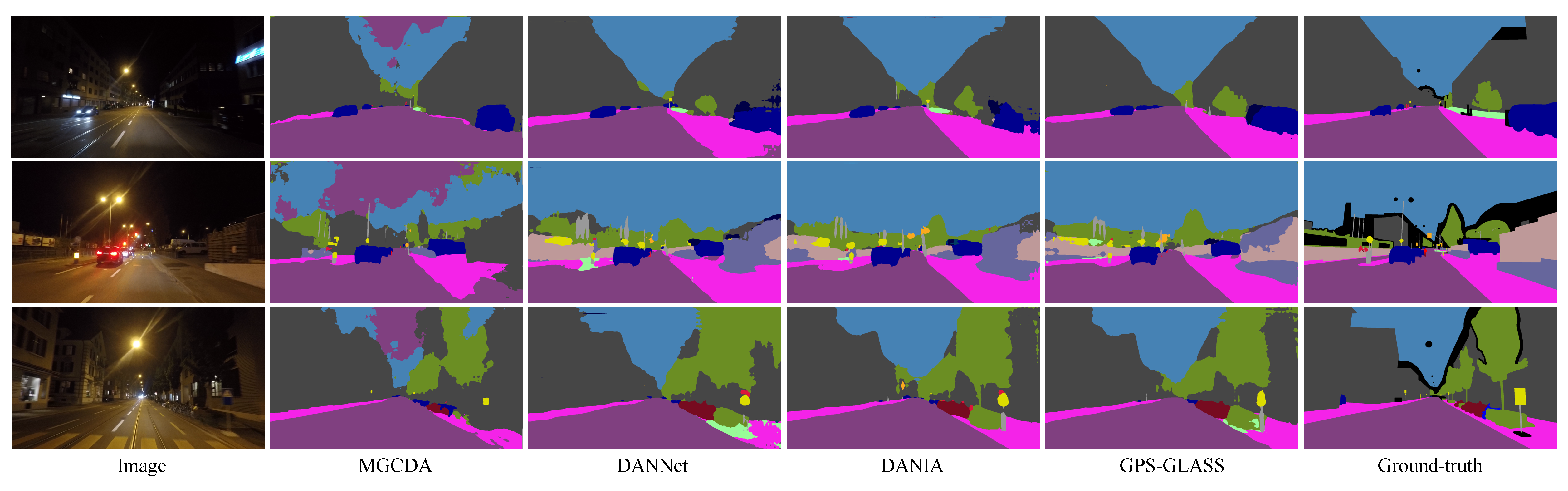}
\caption{Visual comparison of our GPS-GLASS with other state-of-the-art methods on Dark Zurich.}
\label{fig:result}
\end{figure*}

\begin{figure*}[t]
  \centering
  \includegraphics[width=\linewidth]{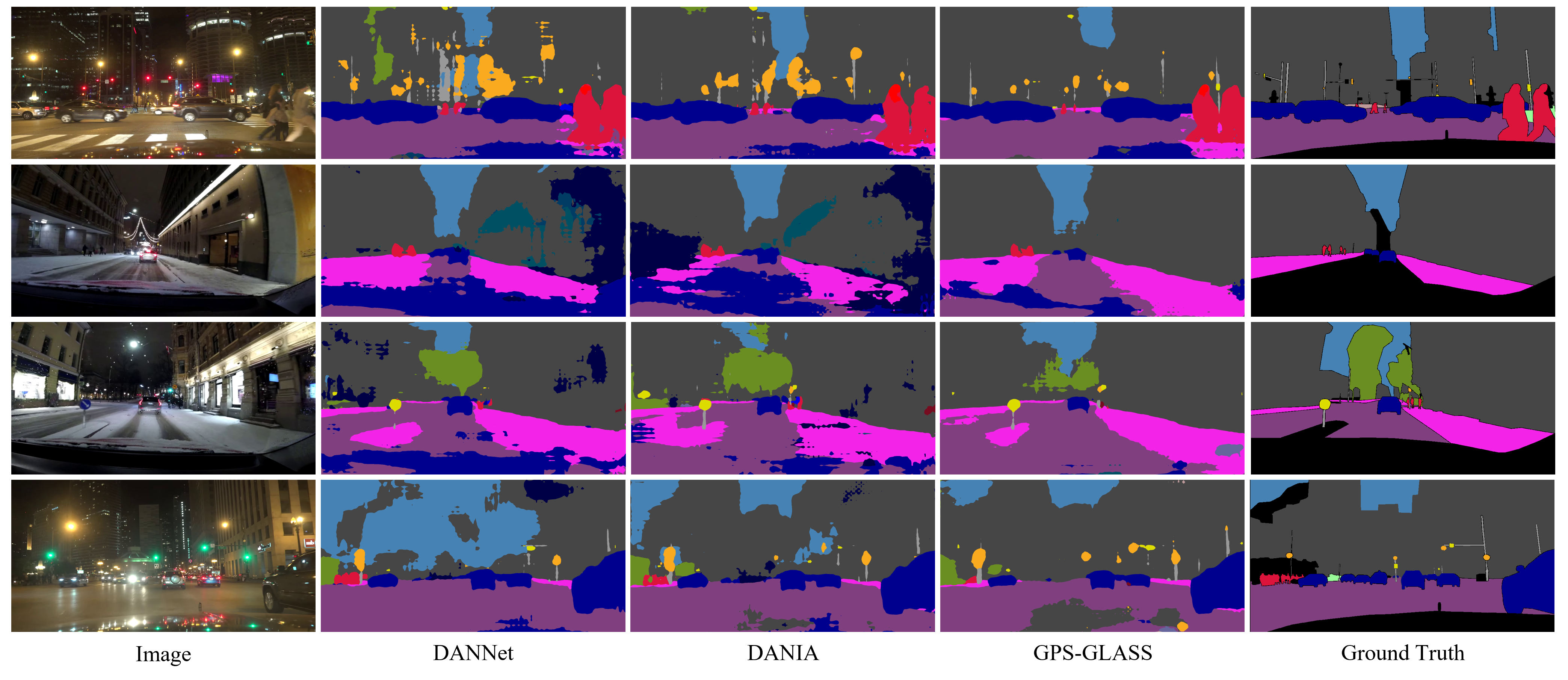}
  \caption{Visual comparison of our GPS-GLASS with other state-of-art methods on NightCity+.}
  \label{figure:nightcity}
\end{figure*}

\begin{figure*}[h!]
\centering
\includegraphics[width=\textwidth]{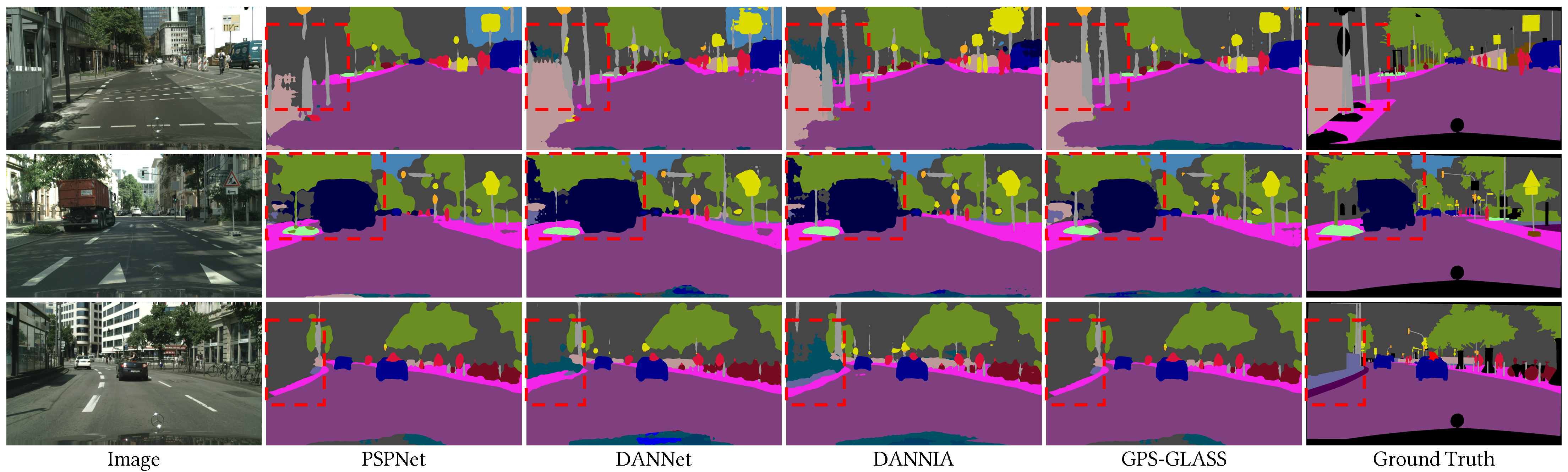}
\caption{Visual comparison of our GPS-GLASS with other state-of-the-art methods on Cityscapes.}
\label{fig:cityscapes}
\end{figure*}

We provide additional results for visual comparison of our proposed GPS-GLASS with state-of-the-art methods. Fig.~\ref{figure:small_scale} shows the results on Dark Zurich-val. Due to pixel-level aligned pseudo-supervision in GPS-GLASS, improvements are especially noticeable in small-scale classes such as fences, poles, and lights. In addition, Figs.~\ref{fig:result} and \ref{figure:nightcity} show the results on Dark Zurich-val and NightCity+, demonstrating the superiority of GPS-GLASS compared to the state-of-the-art methods. In addition, the results on Cityscapes are provided in Fig.~\ref{fig:cityscapes} to demonstrate the generalization ability in the daytime.

%\clearpage
%%%%%%%%% REFERENCES
%{\small
%\bibliographystyle{ieee_fullname}
%\bibliography{egbib}
%}